\documentclass[sensors,article,accept,pdftex,moreauthors]{Definitions/mdpi} 
\firstpage{1} 
\pubvolume{22}
\issuenum{9}
\articlenumber{3109}
\pubyear{2022}
\copyrightyear{2022}
\externaleditor{Academic Editors: {Joaquín Torres-Sospedra, Antoni Perez-Navarro and Raúl Montoliu} 
} 
\datereceived{14 March 2022} 
\dateaccepted{15 April 2022} 
\datepublished{19 April 2022} 
\hreflink{https://\\doi.org/10.3390/s22093109} 

\usepackage{siunitx}
\usepackage{trackchanges}
\addeditor{WJ}
\graphicspath{{./figures/}} 
\Title{Real-Time Sonar Fusion for Layered Navigation Controller $^{\dagger}$}

\TitleCitation{Real-Time Sonar Fusion for Layered Navigation Controller}

\Author{Wouter Jansen 
 $^{1,2,}$*\orcidA{}, Dennis Laurijssen $^{1,2}$\orcidB{} and Jan Steckel $^{1,2}$\orcidC{}}

\AuthorNames{Wouter Jansen, Dennis Laurijssen, Jan Steckel}

\AuthorCitation{Jansen, W.; Laurijssen, D.; Steckel, J.}

\address{%
$^{1}$ \quad Cosys-Lab, Faculty of Applied Engineering, University of Antwerp, 2020 Antwerpen, Belgium;
dennis.laurijssen@uantwerpen.be (D.L.); jan.steckel@uantwerpen.be (J.S.)
\\
$^{2}$ \quad Flanders Make Strategic Research Centre, 3920 Lommel, Belgium}

\corres{Correspondence: wouter.jansen@uantwerpen.be}

\firstnote{\textls[-10]{This paper is an extended version of our paper published in Proceedings of the 2021 International Conference on Indoor Positioning and Indoor Navigation (IPIN), Adaptive Acoustic Flow-Based Navigation with 3D Sonar Sensor Fusion, Lloret de Mar, Spain, 29 November--2 December 2021,~{doi:10.1109/IPIN51156.2021.9662596}.}}

\abstract{Navigation in varied and dynamic indoor environments remains a complex task for autonomous mobile platforms. Especially when conditions worsen, typical sensor modalities may fail to operate optimally and subsequently provide inapt input for safe navigation control. In this study, we present an approach for the navigation of a dynamic indoor environment with a mobile platform with a single or several sonar sensors using a layered control system. These sensors can operate in conditions such as rain, fog, dust, or dirt. The different control layers, such as collision avoidance and corridor following behavior, are activated based on acoustic flow queues in the fusion of the sonar images. The novelty of this work is allowing these sensors to be freely positioned on the mobile platform and providing the framework for designing the optimal navigational outcome based on a zoning system around the mobile platform. Presented in this paper is the acoustic flow model used, as well as the design of the layered controller. Next to validation in simulation, an implementation is presented and validated in a real office environment using a real mobile platform with one, two, or three sonar sensors in real time {with 2D navigation}. Multiple sensor layouts were validated in both the simulation and real experiments to demonstrate that the modular approach for the controller and sensor fusion works optimally. The results of this work show stable and safe navigation of indoor environments with dynamic objects.}


\begin{document}

\section{Introduction}\label{sec:introduction}
This work is an extended version of~\cite{Jansen2021AdaptiveFusion}, where this theoretical foundation was detailed and validated in simulation.
Autonomous navigation of complex and dynamic indoor environments has continued being the subject of extensive research over the last few years. When conditions deteriorate, common sensors such as laser-scanners (LiDAR) and cameras may fail. Concretely, optical sensors could see their medium of structured light distorted by smoke, dust, airborne particles, mist, dirt, or differences in lighting contingent on the sun and cloud conditions. When these types of sensors are used as input for autonomous navigation, object recognition, localization, or mapping systems under such conditions, they may become vulnerabilities for the optimal performance of such systems~\cite{Vargas2021AnConditions}. On~the contrary, other types of sensors, such as radar or sonar, with~their electromagnetic and acoustic sensor modalities, respectively, continue to achieve precise and accurate results, even in rough~conditions. 

Therefore, in-air acoustic sensing with sonar microphone arrays has become an active research topic in the last decade~\cite{Kunin2011DirectionArrays, Verellen2020High-ResolutionArrays, Allevato2022Air-CoupledImaging, Kerstens2019, Allevato2020EmbeddedAcceleration}. This sensing modality can perform well in indoor industrial environments with rough conditions, generate spatial information from that environment, and be used for autonomous activities such as navigation or simultaneous localisation and mapping (SLAM). This type of sonar has a wide field of view (FOV) thanks to the microphone array structure. It can cope with simultaneously arriving echos and transform the recorded microphone signals containing the reflections to full 3D spatial images or point-clouds of the environment. Such biologically inspired~\cite{Griffin1974} sensors have been developed by Jan Steckel~et~al.~\cite{Steckel2020, Kerstens2019} and were implemented for several use-cases, such as for autonomous local navigation~\cite{Steckel2017} and SLAM~\cite{Steckel2013b, Kerstens2020}.

Not only is the sensor modality inspired by nature, but~so is the signal processing. Research on insects that use optical flow clues~\cite{Franceschini2007, Srinivasan2011} and bats using acoustic flow cues~\cite{Corcoran2017,Simon2020,Greiter2017,Warnecke2016} shows that they use these cues to extract the motion of the agent through the environment, also called the ego-motion, and~the spatial 3D structure of said environment through that motion. Jan Steckel~et~al.~\cite{Steckel2017} used acoustic flow cues, found in the images of a sonar sensor, in~a {2D} navigation controller for a mobile platform with motion behaviors such as obstacle avoidance, collision avoidance, and the following of corridors in indoor environments. However, its theoretical foundation had the constraint that only a single sensor could be used, which had to be placed in the center of rotation of the mobile platform. From~a practical point of view, this is not ideal, as mobile platforms in real-world industrial applications will often not suit this constraint of mounting a sensor in that exact position and avoiding 
 an obstructed~FOV. 

Another critical limitation is the limited spatial resolution that using a single sonar sensor can yield compared to the entire environment surrounding the sensor. The~main reasons for this limitation are the leakage of the point-spread function (PSF) and that the sensors have an FOV of \ang{180} in the frontal hemisphere of the sensor, which leaves much of the areas around the sides of a mobile platform undetected. The~PSF leakage may cause undetected reflections. To~solve the described problems, the~most straightforward solution is to use several sonar sensors (multi-sonar) simultaneously and create a spatial resolution that is greater than what the sampling resolution of a single sonar can yield~\cite{Gazit2009}. 

However, digital signal processing of the in-air microphone array data to a 3D spatial image or point-cloud of the environment is a computationally expensive operation. To~deal with this limitation and to support a system such as a mobile platform to use multiple in-air 3D sonar imaging sensors, recent work was published which allows for the synchronization and simultaneous real-time signal processing of a set of sonar sensors on a system with a GPU~\cite{Jansen2020}. That software framework generates a data stream of all the synchronized acoustic (3D) spatial images of the connected sonar sensors. This gives potential for both novel applications, as well as improves existing ones, when compared to only using a single sonar sensor. Specifically on mobile platforms with limited computational resources, this provides new opportunities, as~will become clear in this~paper.

In this work, we will use this gain in spatial resolution and FOV by using a multi-sonar in real time. The~theoretical foundation created in~\cite{Steckel2017} will be expanded to remove the constraint of using one single sonar sensor and to allow a modular design of the mobile platform and placement of sensors. The~navigation controller is expanded to provide the operator the opportunity to create adaptive zoning around the mobile platform where certain motion behaviors should be executed. We expand upon this with a real-time implementation of the navigation controller and additional real experiments in real-world scenarios with sonar sensors. In~these experiments, a real mobile platform navigates an office environment by executing the primitive motion behaviors of the~controller. 

We will describe the 3D sonar sensor in Section~\ref{sec:ertis}. Subsequently, we will be detailing the acoustic flow paradigm in Section~\ref{sec:acoustic_flow_model} and its implementation in the {2D} navigation controller in Section~\ref{sec:control}. Afterwards, we will provide validation and discussion of the layered controller with simulation and real-world experiments in Section~\ref{sec:results}. Finally, we will conclude in Section~\ref{sec:conclusion} and briefly look at the future of this research.

\section{Real-Time 3D Imaging~Sonar}\label{sec:ertis}
\begin{figure}[H]
\begin{adjustwidth}{-\extralength}{0cm}
\centering 
\begin{tabular}{ccc}
\includegraphics[width=0.28\textwidth]{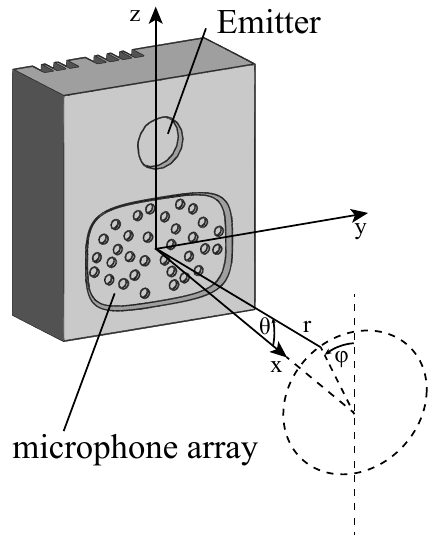}&\includegraphics[width=0.35\textwidth]{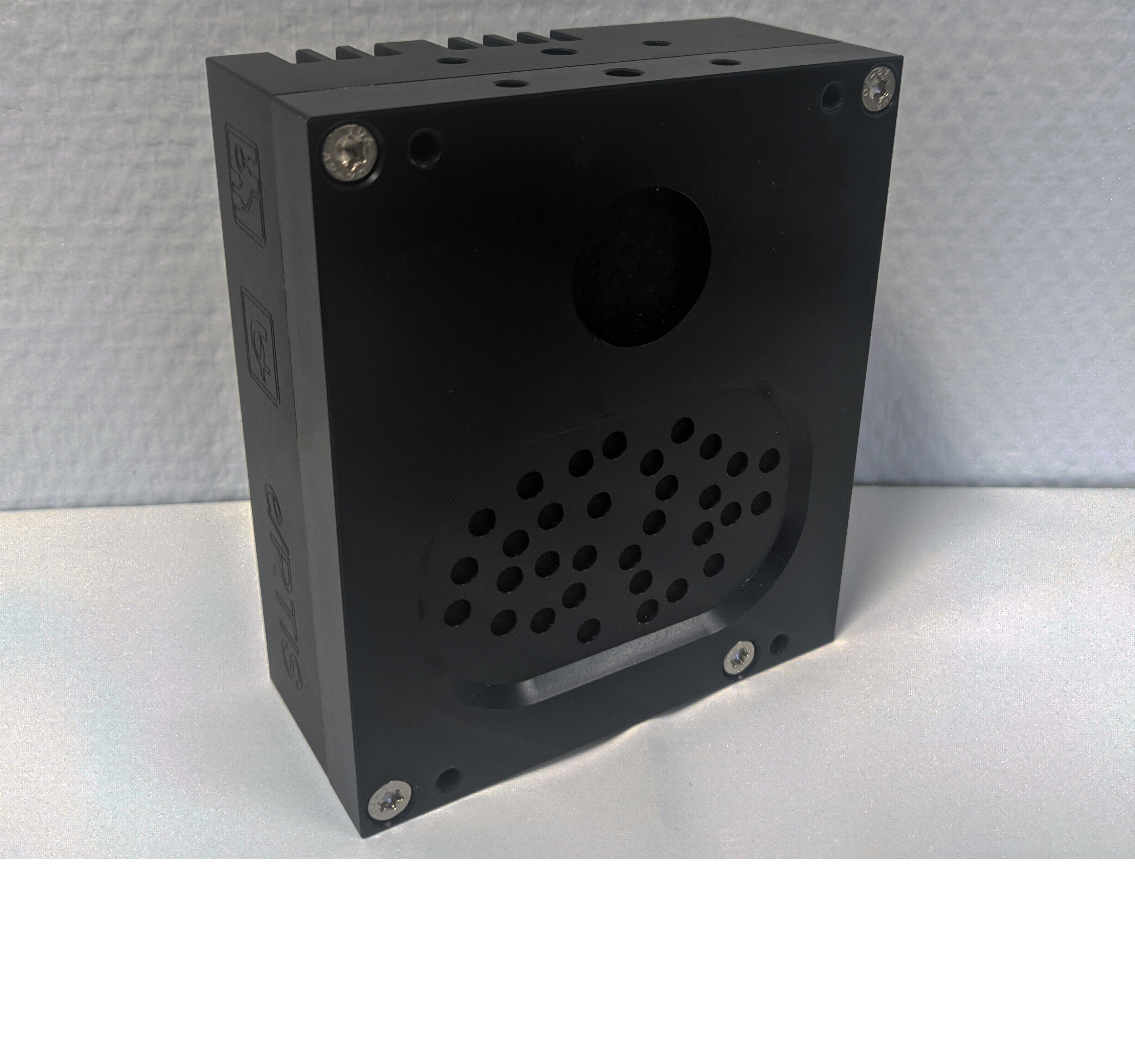}&
\includegraphics[width=0.35\textwidth]{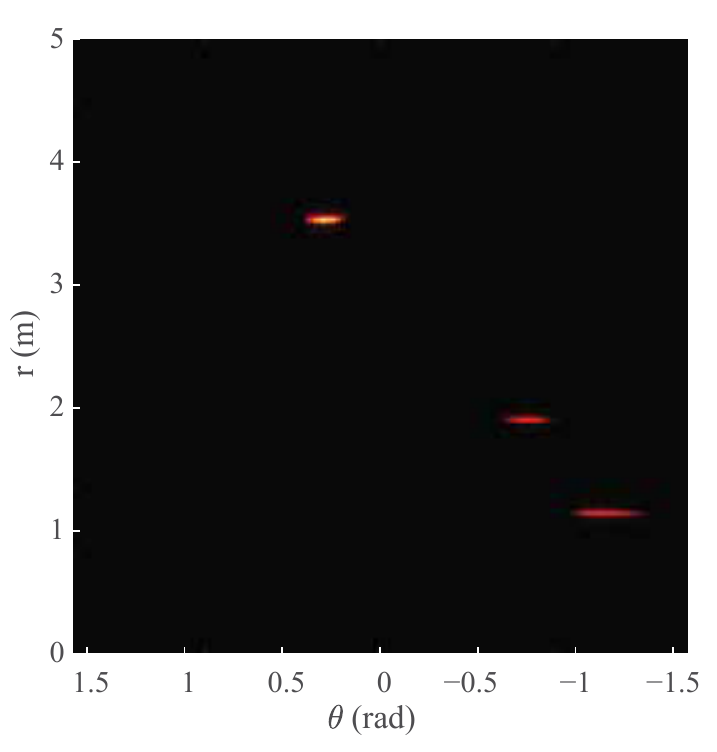}\\
(\textbf{a})&(\textbf{b})&(\textbf{c})\\
\end{tabular}  
\end{adjustwidth}
  \caption{(\textbf{a}) The location of a reflection expressed in the spherical coordinate system ($r,\theta,\varphi$) associated with the sensor. The~embedded real-time imaging sonar (eRTIS) sensor is drawn to show the irregularly positioned array of 32 digital microphones and the emitter. (\textbf{b}) A photo taken from the eRTIS sensor, as used in the real experiments. (\textbf{c}) An example of a 2D energyscape, a~2D sonar image of an \ang{180} horizontal scan. {Subfigures (\textbf{b},\textbf{c}) were added for better visualisation of the sensor and energyscape concepts. 
}}
  \label{fig:sensor_images}
\end{figure}

\begin{figure}[H]
\includegraphics[width=0.98\linewidth]{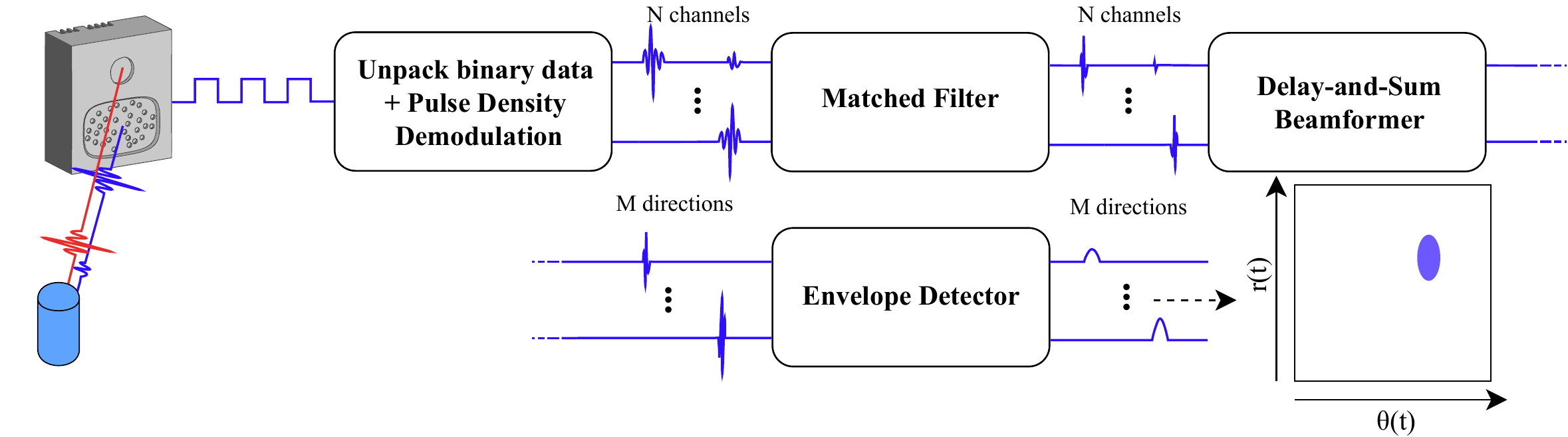}
\caption{Diagram of the digital signal processing steps to go from the raw microphone signals to the energyscapes. The~initial signals containing the reflections of the detected objects are first demodulated to the full audio signals of the N channels. A~matched filter is used for finding the actual reflections of the emitted signal. Delay-and-sum beamforming creates a spacial filter for every direction of interest. Subsequently, envelope detection is used to clean up the spatial~image.}
\label{fig:ertis_processing}
\end{figure}

First, we will provide a synopsis of the 3D imaging sonar sensor used, the~embedded real-time imaging sonar (eRTIS) \cite{Steckel2020}. A~much more detailed description can be found \mbox{in~\cite{Kerstens2019, Kerstens2020}.} The~sensor, drawn in Figure~\ref{fig:sensor_images}a and photographed in Figure~\ref{fig:sensor_images}b, contains a single emitter and uses a known, irregularly positioned array of 32 digital micro-electromechanical systems (MEMS) microphones. A~broadband FM-sweep is sent out by the the ultrasonic emitter. The~subsequent acoustic reflections will be captured by the microphone array when this FM-sweep is reflected on objects in the environment. Using these specific chirps in the FM-sweep, we can increase the spatial resolution~\cite{Steckel2013} of a single sonar sensor. Thereupon, the~spatial images are created from digital signal processing of the microphone signals recorded by the sensors. Such a spatial image, which we also call an energyscape, uses a spherical coordinate system with ($r,\theta,\varphi$). This coordinate system is shown in Figure~\ref{fig:sensor_images}a, while an example of an energyscape is shown in Figure~\ref{fig:sensor_images}c. A~brief overview of the processing pipeline is shown in Figure~\ref{fig:ertis_processing}. The~current eRTIS sensors have an FOV of \ang{180} for the vertical and horizontal~FOVs. \vspace{-6pt}

\vspace{-9pt} 
When multiple eRTIS sensors are simultaneously used with overlapping FOVs, their emitters are synchronized up to 400 ns \cite{Kerstens2020}. 
 to avoid the unwanted artifacts by picking up the FM-sweep of another eRTIS sensor that was sent out at an other position in space and time. Furthermore, artifacts caused by multi-path reflections of other sensors do not cause issues in the results of this paper, as these types of reflections would only cause an inaccuracy in the resulting range and not angle. By~using the sensor fusion model detailed further in this paper, where only the closest reflections for each angle would cause actuation of certain motion behavior, the~resulting reflections would not cause 
  any unwanted behavior.

Recently, a~software framework was created that can handle multiple eRTIS sensors and perform the signal processing in real time using GPU-acceleration~\cite{Jansen2020}. Furthermore, the~latest generation of eRTIS sensors can incorporate an NVIDIA Jetson module embedded into the casing such that the signal processing can be performed on board. A~mobile platform equipped with multiple eRTIS sensors with embedded NVIDIA Jetson TX2 NX modules is able to process their sonar images in real time. A~measurement frequency of \SI{10}{\hertz} is used in the experiments of this paper. 


\section{Acoustic Flow~Model}\label{sec:acoustic_flow_model}
Energyscapes of an eRTIS device contain the coordinates ($r,\theta,\varphi$) of a reflection in the seen environment. The~theoretical acoustic flow model described here consists of a differential equation of the spatial 3D location of the object represented by this reflection with the rotation and linear ego-motions of the mobile platform. In~other words, if~the mobile platform moves, the~model describes how the static reflection will be moving within the~energyscape.

\begin{figure}[H]
\includegraphics[width=0.36\linewidth]{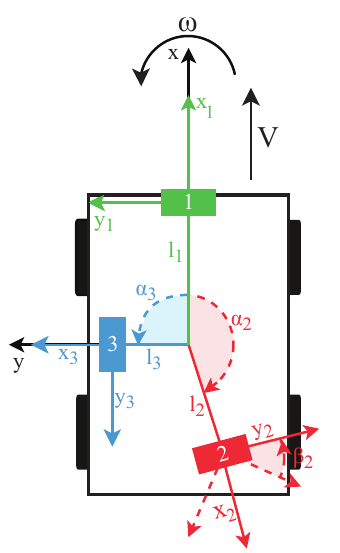}
\caption{Drawing of a mobile platform seen from the top with three eRTIS sensors. Each is described by the distance $l$ from the platform's rotation center point, the~angle $\alpha$ around on the XY-plane relative to the x-axis, and the angle $\beta$ which describes the local rotation of the sensor around its own center. The~only constraint is that all sensors must be located roughly on the same horizontal~XY-plane.}
\label{fig:robot_coordinates}
\end{figure}

\textls[-15]{This model is based on the work in~\cite{Steckel2017}. However, it only works for a singular sonar sensor that faces forward. Therefore, the~model is adjusted in this paper to allow multi-sonar, where several sensors can be placed on the mobile platform with the only constraint being that they must all be positioned on the same horizontal plane. Figure~\ref{fig:robot_coordinates} shows a schematic of a multi-sonar~configuration.
The model described here is the same as in the original paper~\cite{Jansen2021AdaptiveFusion}, but is detailed here once more to provide all necessary context to \mbox{the reader.}}
\vspace{-6pt} 

\subsection{2D-Velocity~Field}
A vector $(r,\theta,\varphi)$ is used for defining the location of a reflector in the sonar sensor reference frame, as shown in Figure~\ref{fig:sensor_images}a. The~same spherical coordinate system is used within the energyscapes of the eRTIS devices.
This spherical coordinate system is related to the standard right-handed Cartesian coordinates by
\begin{equation}
\vec{x}(t) =
\begin{bmatrix}
    x(t)  \\
    y(t)  \\
    z(t)
\end{bmatrix}=
\begin{bmatrix}
    r(t)\cos(\theta(t))  \\
    -r(t)\sin(\theta(t))\sin(\varphi(t))  \\
    r(t)\sin(\theta(t))\cos(\varphi(t))
\end{bmatrix}
\label{eq:af_coordinate_system}
\end{equation}

Expressed in the reference frame of the sensor, the~time derivatives of the stationary reflector are given by
\begin{equation}
\frac{d\vec{x}}{dt} = -\vec{v}_{s} - \vec{\omega}_{s}\times\vec{x}
\label{eq:af_reflector_time_derivative}
\end{equation}
with symbols $\vec{v}_{s}$ and $\vec{\omega}_{s}$ representing the sensor's linear and rotational velocity vectors, respectively, in~the reference frame of the~sensor.

The x-axis of the mobile platform's frame, as shown in Figure~\ref{fig:robot_coordinates}, is defined as the direction of the linear velocity vector. Furthermore, the~mobile platform can only perform rotations about its z-axis. With~these constraints, the~earlier-defined velocity vectors of the sensor are
\begin{equation}
\vec{v_{s}} =
\begin{bmatrix}
    V\cos(\alpha + \beta) + \sin(\beta) l  \omega \\
    V\sin(\alpha + \beta) + \cos(\beta) l \omega\\
    0
\end{bmatrix},
\vec{\omega_{s}} =
\begin{bmatrix}
    0  \\
    0  \\
    \omega
\end{bmatrix} 
\label{eq:af_velocities}
\end{equation}
\begin{equation}
\delta = \alpha + \beta
\label{eq:delta_ef}
\end{equation}
with $V$ and $\omega$ representing the magnitudes of the linear and angular velocities of the mobile platform, respectively, which are constant during a single measurement. To~describe the location of a sensor from the reference frame of the mobile platform, the~parameters ($l,\alpha,\beta$) are used. $\alpha$ describes the rotation around the z-axis of the mobile-platform relative to the x-axis, and $\beta$ defines the rotation on the sensor around its own center point. Finally, $l$ describes the distance between the sensor and the center of rotation of the mobile platform. These parameters are also shown in Figure~\ref{fig:robot_coordinates}. Taking the derivative of Equation, we obtain~\eqref{eq:af_coordinate_system} 
\begin{adjustwidth}{-\extralength}{0cm}
\begin{equation}
\frac{d\vec{x}}{dt} = 
\begin{bmatrix}
    \cos(\theta(t)) & -r(t)\sin(\theta(t)) & 0 \\
    -\sin(\theta(t)) \sin(\varphi(t)) & -r(t)\cos(\theta(t)) \sin(\varphi(t)) & -r(t)\sin(\theta(t)) \cos(\varphi(t))\\
    \sin(\theta(t)) \cos(\varphi(t)) & r(t)\cos(\theta(t)) \cos(\varphi(t)) & -r(t)\sin(\theta(t)) \sin(\varphi(t))
\end{bmatrix}\begin{bmatrix}
    dr/dt  \\
    d\theta/dt  \\
    d\varphi/dt
\end{bmatrix} 
\label{eq:af_coordinate_derivative}
\end{equation}
\end{adjustwidth}

Afterwards, Equations \eqref{eq:af_coordinate_system}, \eqref{eq:af_velocities} and \eqref{eq:af_coordinate_derivative} are placed in Equation \eqref{eq:af_reflector_time_derivative}. Furthermore, as~we use the earlier-described constraint of having all sensors on the same horizontal plane and are performing only 2D planar navigation, the~reflectors are restricted to lie in the horizontal plane, i.e.,~$\varphi = \{-\pi/2,\pi/2\}$. With~this simplification and applying trigonometric sum and difference identities, the~2D-velocity field can be expressed as
\begin{equation}
\begin{bmatrix}
dr/dt\\
d\theta/dt
\end{bmatrix}=\begin{bmatrix}
    -l\omega \sin(\beta - \theta(t)) -V\cos(\theta(t) + \delta)\\
   \cfrac{l\omega \cos(\beta - \theta(t)) + V\sin(\theta(t) + \delta)}{r(t)} + \omega
\end{bmatrix} 
\label{eq:af_refl_ego_motion}
\end{equation}

\subsection{Only Linear~Motion}
If the linear motion of the mobile platform is limited, i.e.,~$\omega$ = 0 rad/sec, the~vehicle can only move forward or backwards. In~this case, Equation \eqref{eq:af_refl_ego_motion} can be reduced to find the time derivatives of the reflector coordinates to make a 2D acoustic flow model for \mbox{linear motion:}
\begin{equation}
\begin{gathered}
dr/dt = -V\cos(\theta(t) + \delta) \\
d\theta/dt = \cfrac{V}{r(t)}\sin(\theta(t) + \delta)
\end{gathered}
\label{eq:af_diff_refl_linear}
\end{equation}

The solution to these differential equations has to comply with
\begin{equation}
\cfrac{\dot{r}(t)}{r(t)} = - \dot{\theta}(t)\cfrac{\cos(\theta(t) + \delta)}{\sin(\theta(t) + \delta)}
\label{eq:af_diff_refl_linear_comply}
\end{equation}
with $\theta_0$ describing the angle between the eRTIS sensor and the first sighting of the reflector and not being equal to 0. Subsequently, integrating both sides of Equation \eqref{eq:af_diff_refl_linear_comply}, we can find a constant $C$, defined as
\begin{equation}
C = |r(t)|  \sin(\theta(t) +\delta)
\label{eq:af_diff_refl_linear_solution}
\end{equation}

\begin{figure}[H]
\begin{tabular}{cc}
\includegraphics[width=0.53\textwidth]{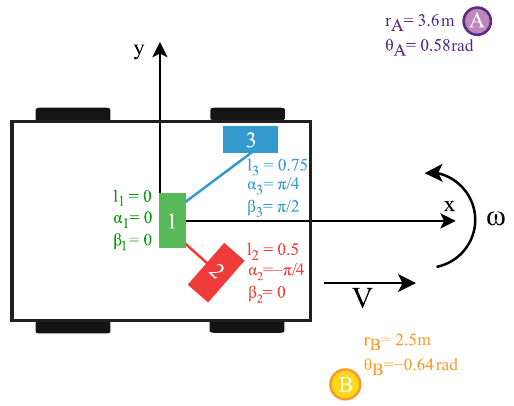}\\
(\textbf{a})\\
\includegraphics[width=0.95\textwidth]{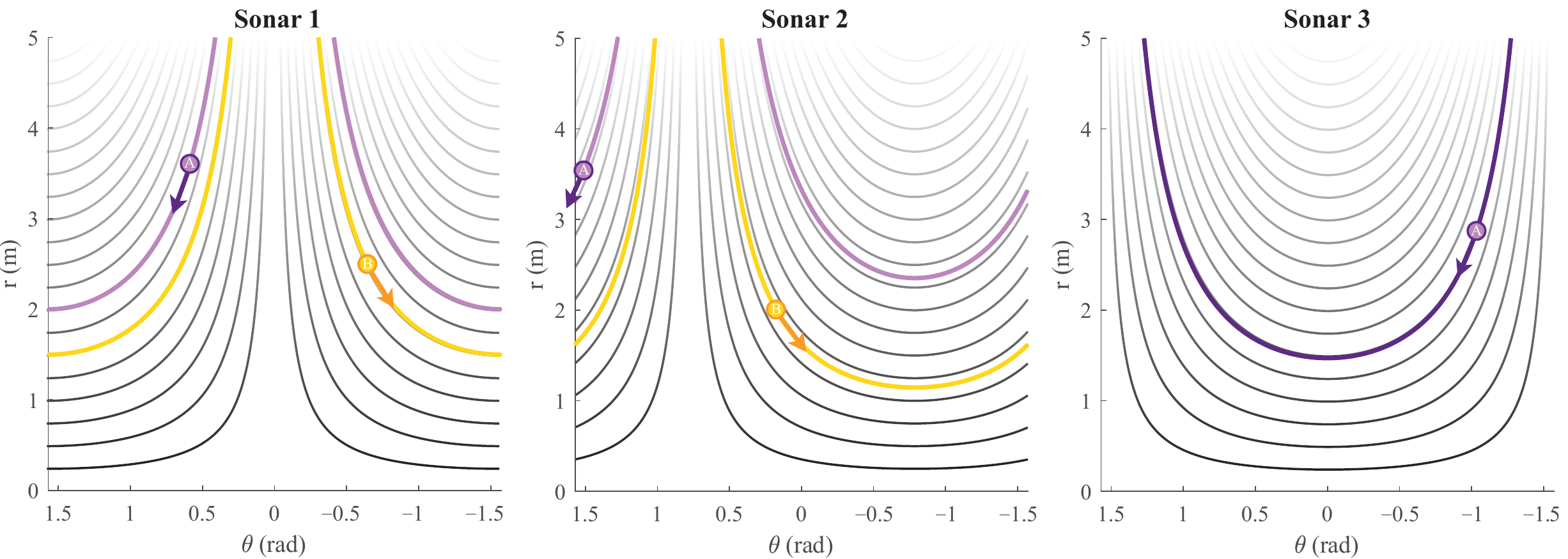}\\
(\textbf{b})\\
  \end{tabular}
  \caption{(\textbf{a}) A mobile platform seen from the top with three sonar sensors. The~$(\alpha, \beta$, $l)$ parameter values are shown as well. Next to it, two reflecting objects exist, defined as A and B with their $(r_0, \theta_0$) coordinates. {The placement of these reflectors in the figure is not accurate or to scale.} (\textbf{b})~The $(r(t), \theta(t))$ flow-lines of reflections A and B for a positive linear motion $V$ are marked for sonar sensors 1, 2, and 3 of Figure~\ref{fig:af_flow_lines_linear}a. The~arrow indicates the direction the reflection would move with such a motion. The~flow-lines for other distances falling in range ($r \in [0; r_{max}]$) are also shown, with~$r_{max}$ being the maximum detected range of the eRTIS sensor. 
}
  \label{fig:af_flow_lines_linear}
\end{figure}

This defines the solutions of the differential Equation \eqref{eq:af_diff_refl_linear} for chosen starting conditions $(r_0, \theta_0)$. These are the different flow-lines for all chosen ranges, as shown in Figure~\ref{fig:af_flow_lines_linear}b.\vspace{-6pt}

\subsection{Only Rotation~Motion}
If we limit the mobile platform to a pure rotation, i.e.,~$V$ = 0 m/sec, the time derivatives of the coordinates $(r(t), \theta(t))$ of a reflection of Equation \eqref{eq:af_refl_ego_motion} reduce to
\begin{equation}
\begin{gathered}
dr/dt = -l\omega  \sin(\beta - \theta(t))\\
d\theta/dt = \cfrac{l\omega}{r(t)}  \cos(\beta - \theta(t)) + \omega
\end{gathered}
\label{eq:af_diff_refl_rotation}
\end{equation}

\begin{figure}[H]
\includegraphics[width=0.95\linewidth]{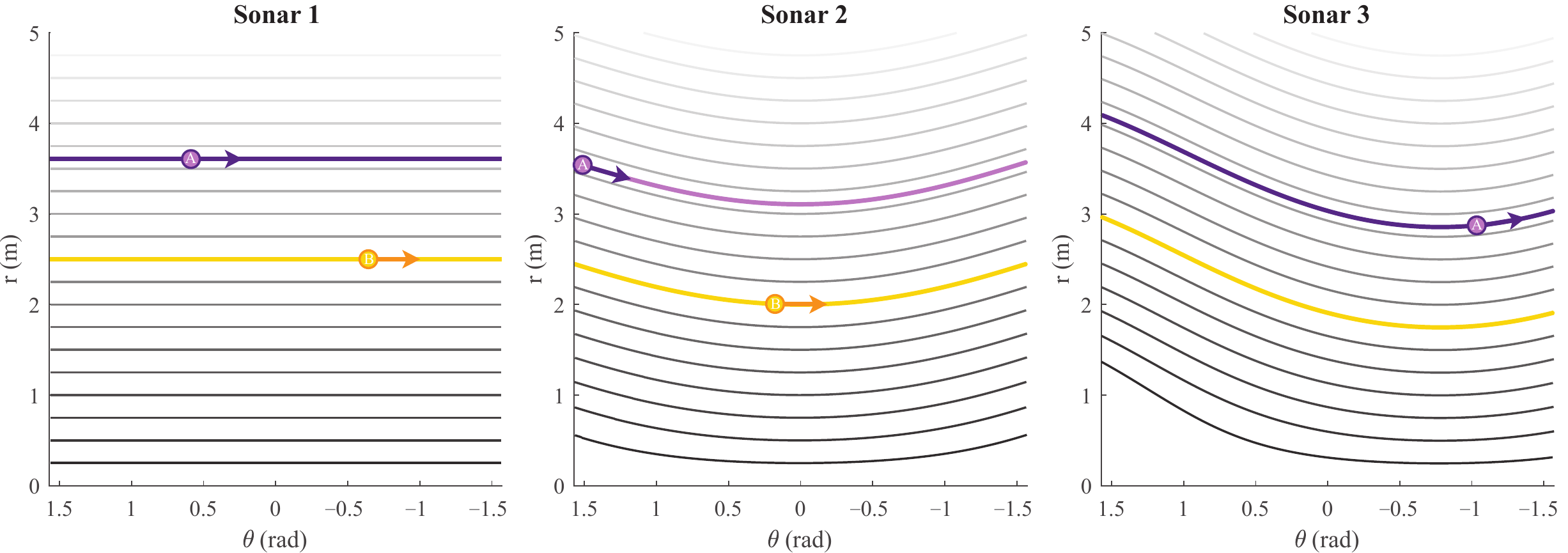}
\caption{The $(r(t),\theta(t))$ flow-lines for a positive rotation movement $\omega$ are marked for sonar sensors 1, 2, and 3 and reflections A and B of Figure~\ref{fig:af_flow_lines_linear}a. The~arrow indicates the direction the reflection would move with such a motion. The~flow-lines for other distances falling in range ($r \in [0; r_{max}]$) are also shown, with~$r_{max}$ being the maximum detected range of the eRTIS sensor. 
}
\label{fig:af_flow_lines_rotation_graph}
\end{figure}

These flow-lines are shown in Figure~\ref{fig:af_flow_lines_rotation_graph}. A~closed-form expression, similar to that for linear motion, could not be found for the case of a rotation~motion.

\section{Layered Navigation~Controller}\label{sec:control}
\begin{figure}[H]
\includegraphics[width=1\linewidth]{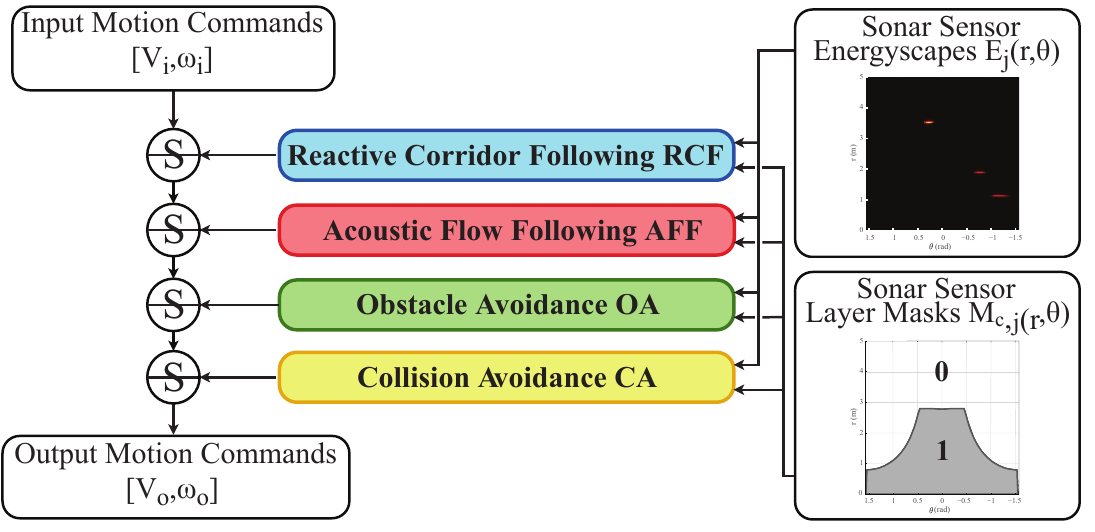}
\caption{Diagram of the subsumption architecture of the controller. The~input of each layer is the energyscape and acoustic mask where $c$ stands for the active motion primitive layer and $j$ defines the index of the sensor. Note that the lower the layer is in the list, the~higher its priority will be in a subsumption architecture. 
}
\label{fig:control_layers}
\end{figure}

With the acoustic flow model formulated in Section~\ref{sec:acoustic_flow_model}, we can know a priori how a reflection will move through the energyscape based on the linear and rotational velocity of the mobile platform, and subsequently execute motion commands based on this new a priori knowledge. We suggest using acoustic regional zoning around the mobile platform in such a way that when the reflection of an object or obstacle is perceived in one of these predefined zones, certain primitive motion behaviors are activated. 
A layered control system designed as a subsumption architecture, first introduced by R. A. Brooks~\cite{Brooks1986}, is used. A~ diagram of the layered navigation controller is shown in Figure~\ref{fig:control_layers} combined with its input and output variables. The~inputs are the linear and rotational velocities $[V_i,\omega_i]$. These can originate from several other systems, such as, for example, a manual human operator or a global path planner. The~output velocities $[V_o,\omega_o]$ of the layered controller is the modulation of these input velocities by the behavior law of one of the layers. These layers themselves also require input, which comes in the form of the sonar energyscapes of each eRTIS sensor, as well as the mask for that sensor and layer, which will be explained further in this section. The~individual control behaviors are based on~\cite{Steckel2017} but have been adapted in this work to be stable and configurable for multi-sonar setups and to support the new acoustic regional zoning~system.

\subsection{Acoustic Control Regions and~Masks}
\begin{figure}[H]
\includegraphics[width=0.5\linewidth]{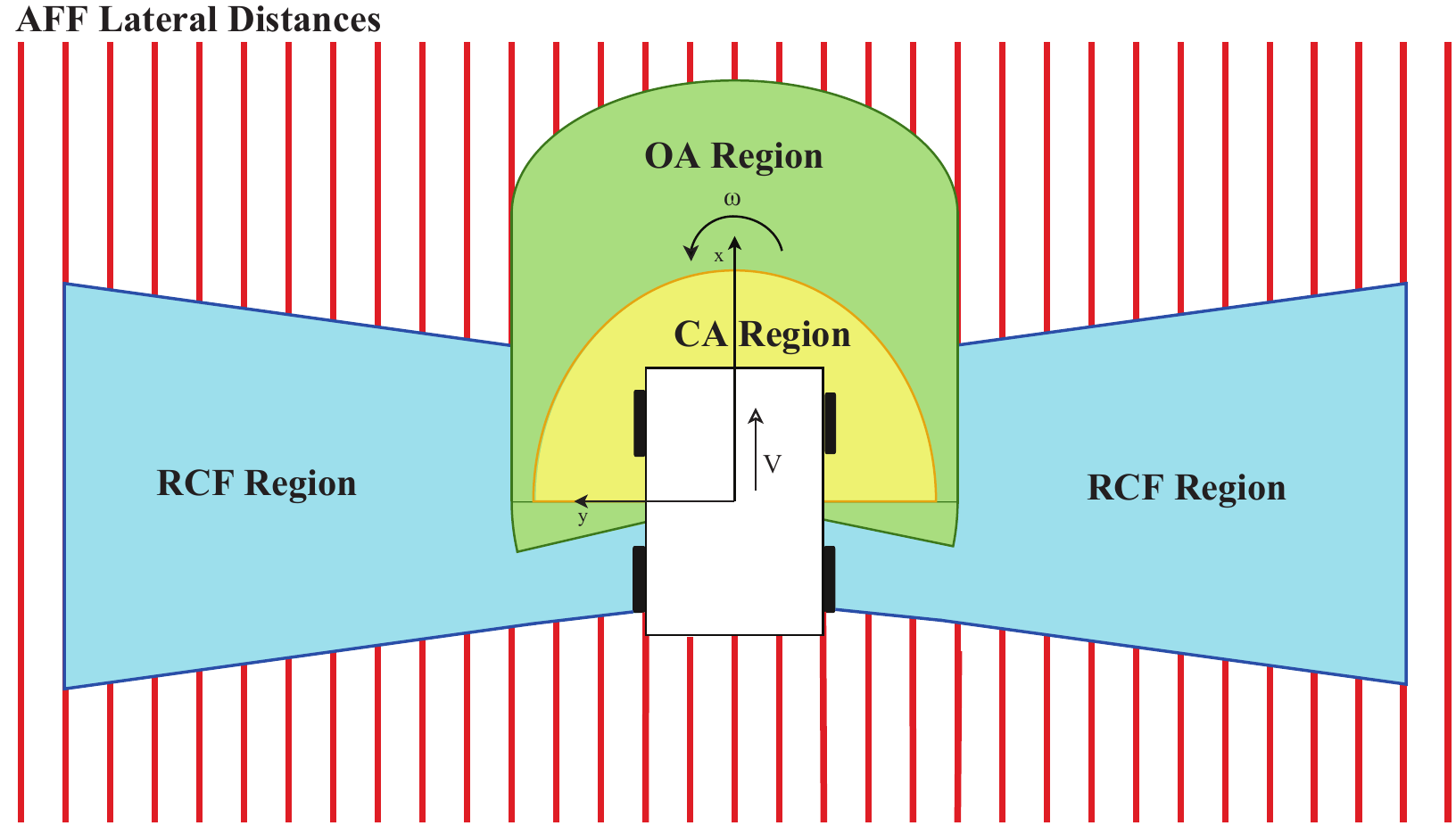}
\caption{A region is defined for every control layer around the mobile platform. They are described by lines parallel to the mobile platform's x-axis, a~circular shape with a certain radius from the center of the platform or limited by an angle $\theta$ from that same center point. The~described shapes provide the input for the acoustic flow models to generate masks within the eneryscapes of each sonar~sensor.}
\label{fig:control_robot_regions}
\end{figure}

\begin{figure}[H]
\includegraphics[width=0.95\linewidth]{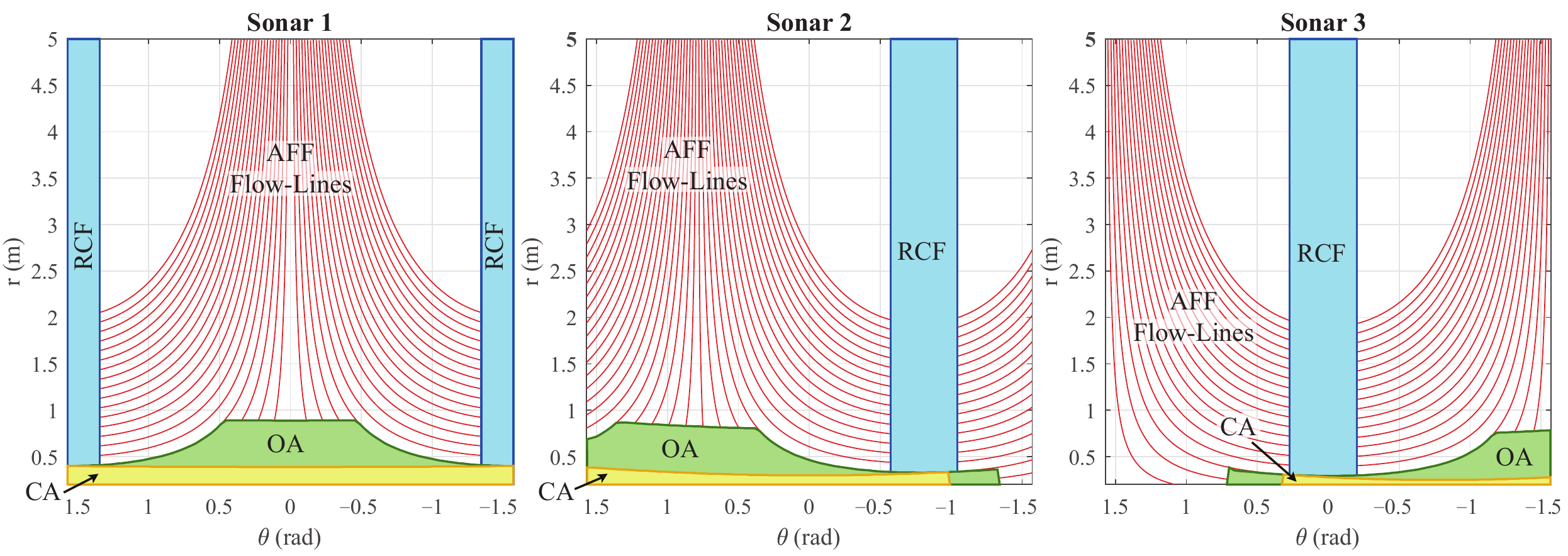}
\caption{The generated masks for each eRTIS sensor in Figure~\ref{fig:af_flow_lines_linear}a are drawn for the control zones of Figure~\ref{fig:control_robot_regions} for the four different motion behavior layers. 
}
\label{fig:control_masks}
\end{figure}

A zone is defined around the mobile platform for each layer. An~example of the zone for each layer is shown in Figure~\ref{fig:control_robot_regions}. Such a zone can be defined by basic contours such as straight lines, rectangles, circles, or trapezoids. These zones are constrained to these primitive shapes according to the time-differential equations of Section~\ref{sec:acoustic_flow_model}. The~flow-lines these equations provide are used as the outer edges of these contours within the area that can be perceived by the sensor in its 2D energyscape. A~mask $M_{c,j}$ with $c$ the behavior layer and the index $j$ of the sonar sensor can be made for each layer and sensor pairing. We created this mask as a matrix with the same resolution as the energyscape. Figure~\ref{fig:control_masks} illustrates some examples of such acoustic masks based on the acoustic flow~model.

If we cover an energyscape $E_j$ with a mask, and a certain amount of reflection energy is located in the the remaining active area, i.e.,~not nullified by the mask, a~behavior layer can be excited if this abides by its control law with certain constraints, such as being above a certain threshold. These constraints will be described further when each layer is separately discussed later. A~second feature of the masking system is to allow the controller to indicate if a reflector is on either the left or right side of the mobile platform. This creates the possibility for a behavior layer to decide on the correct steering direction for the output velocity commands. We use ternary masks, which make a voxel of the mask 0 when it is not within the active area of the mask, 1 if it is on the left side, and $-$1 if it is on the right side of the mobile platform,~respectively.

The design of the navigation controller described here is the same as in the original paper~\cite{Jansen2021AdaptiveFusion} but is detailed here once more to provide all necessary context to the reader.
\unskip
\unskip
\subsection{Collision~Avoidance}
The lowest layer is collision avoidance (CA), which has the highest priority due to the subsumption architecture. This behavior layer is triggered if a certain amount of reflection energy above the threshold $T_{CA}$ is perceived within the CA region. This region, of~which an example is illustrated in Figure~\ref{fig:control_robot_regions}, is used to generate a mask for each eRTIS sensor $M_{CA,j}$, as shown in Figure~\ref{fig:control_masks}. We use a half-circle near the front of the mobile platform so that when steering along a path, collision can be avoided using this~region.

The collision avoidance control law $C_{CA}$ is
\begin{equation}
C_{CA} : \exists r,\theta: \sum_{r,\theta,j} E_j(r,\theta) M_{CA,j}(r,\theta) > T_{CA}
\label{eq:ca_law}
\end{equation}

This results in an opposite motion of the nearest reflection. This motion is the rotational output velocity $\omega_o$ and is set as a constant for which we used \SI{0.5}{\radian/\second} in our experiments. To~know the sign of the output rotational velocity, the ternary masking as described earlier is~used.

The linear output velocity $V_o$ for the mobile platform is always \SI{0}{\meter/\second} until this layer is triggered four times or more in succession. Once this occurs, a~small reverse motion is started, with, in our experiments, this linear velocity being set to \SI{-0.1}{\meter/\second}. These output velocities are sustained as long as there is reflection energy in the region of the CA layer above the threshold $T_{CA}$. 
\subsection{Obstacle~Avoidance}
Obstacle avoidance (OA) is the the subsequent layer in terms of priority. Its behavior is triggered when the total amount of reflection energy in multiplication of the mask $M_{OA,j}$ and the energyscape $E_j$ is above the threshold $T_{OA}$. An~example of this region is shown in Figure~\ref{fig:control_robot_regions}. It is designed in a way that as long as a forward linear velocity $V$ is sustained without a rotational component ($\omega=0$), a~collision is likely to occur. Therefore, the~control law $C_{OA}$ is designed to circumvent this imminent collision. It is defined as
\begin{equation}
\begin{split}
& C_{OA} : \exists r,\theta: E_j(r,\theta) M_{OA,j}(r,\theta) > T_{OA}\\
& \omega_{OA} = \omega_i + \lambda_{OA} \cfrac{\sum_{r,\theta,j} \frac{1}{r^2}  E_j(r,\theta) M_{OA,j}(r,\theta)}{\sum_{r,\theta,j}E_j(r,\theta) M_{OA,j}(r,\theta)}\\
& V_{OA} = V_i\left[1-\mu_{OA} \sum_{r,\theta,j}E_j(r,\theta) M_{OA,j}(r,\theta)\right]
\end{split}
\label{eq:oa_law}
\end{equation}

The outputs are $V_o=V_{OA}$ and $\omega_o=\omega_{OA}$. In~practice, this will result in the mobile platform moving farther away from the side with the largest total reflection within the masked energyscape summation. This is possible thanks to the ternary masks. As~can be observed in Equation \eqref{eq:oa_law}, the~division of the masked energyscapes by the square of the range of active voxels containing reflection energy, the~further away this reflection energy is, 
 the more it will result in a smaller avoidance motion. Subsequently, the~avoidance behavior will slowly ramp up towards the obstacle to avoid unwanted drastic movement. Additionally, the~linear output velocity will be lowered based on the range of the total energy in order to slow down the mobile platform as the obstacle moves closer. Gain factors to further tweak the output velocities $\lambda_{OA}$ and $\mu_{OA}$ were both set to 1 in simulation and 0.1 for the real experiments, but can be modified.
\subsection{Reactive Corridor~Following}
We will now discuss the last layer of the subsumption architecture shown in Figure~\ref{fig:control_layers}, reactive corridor following (RCF), as it establishes the initial alignment with a detected corridor quickly. By~giving this layer the last position, acoustic flow following (AFF) will be higher in priority and activate once this initial alignment is achieved. This alignment is created by balancing the reflection energy in the summation of the masked energyscapes of the left and right peripheral regions of the mobile platform. Simply put, the~control law will cause the platform to move away from the side with the most energy. These peripheral regions are shown in the example illustrated in Figure~\ref{fig:control_robot_regions}. The~masks generated from this region are shown in Figure~\ref{fig:control_masks}. This layer will be triggered when the total reflection energy is larger than the threshold $T_{RCF}$.
The law $C_{RCF}$ is expressed as
\begin{equation}
\begin{split}
& C_{RCF} : \exists r,\theta: E_j(r,\theta) M_{RCF,j}(r,\theta) > T_{RCF}\\
& \omega_{RCF} = \omega_i + \lambda_{RCF} \cfrac{\sum_{r,\theta,j} \frac{1}{r^2}  E_j(r,\theta) M_{RCF,j}(r,\theta)}{\sum_{r,\theta,j}E_j(r,\theta) M_{RCF,j}(r,\theta)}\\
& V_{RCF} = V_i
\end{split}
\label{eq:rcf_law}
\end{equation}

The output motion of this layer is the velocities $V_o=V_{RCF}=V_i$ and $\omega_o=\omega_{RCF}$ which are dependent on a gain factor $\lambda_{RCF}$. It is set to 1 in simulation and 0.1 for the \mbox{real experiments.}
\subsection{Acoustic Flow~Following}
If the RCF layer has achieved alignment with a corridor, or when the mobile platform arrives near one or two parallel walls, the~AFF layer is triggered. In such an event, the~detected reflections on the wall(s) can be found on the equivalent flow-line for the distance to that wall, as described in Section~\ref{sec:acoustic_flow_model} and on Figure~\ref{fig:af_flow_lines_linear}b. This alignment for a sensor $j$ can be measured with
\begin{equation}
\Gamma_j(d) = \cfrac{\sum_{r,\theta}E_j(r,\theta) M_{AFF,F_{d,j}}(r,\theta)   \sqrt[\leftroot{0}\uproot{3}]{r}}{|F_{d,j}|}
\label{eq:AF_integral}
\end{equation}

In principle, this formula is integrating all voxels on a flow-line at the lateral distance $d$ from the path of the mobile platform. Several lateral distances are shown in Figure~\ref{fig:control_robot_regions}, with the associated flow-lines visualized in Figure~\ref{fig:control_masks}.  In~Equation \eqref{eq:AF_integral}, $F_{d,j}$ is representing the flow-line at the lateral distance $d$, and $|F_{d,j}|$ denotes its length, i.e.,~the amount of voxels that lay on that flow-line.
For every combination of sensor $j$ and distance $d$, $F_{d,j}$ is generated. 
Fusing of all the $F_{d,j}$ values gives the total alignment for that lateral distance $F_{d}$. When one such value peaks and is greater than the threshold $T_{AF,single}$, this layer becomes actuated. If~only one peak is detected, motion behavior will be started to continue to follow the wall at lateral distance $d_{s}$ associated with that peak. This behavior is defined as
\begin{equation}
\begin{split}
& \omega_{AFF} = 
\begin{cases}
      \omega_i + \lambda_{AFF} ( d_s - d_p) & \text{if } d_p \text{ is defined}\\
      \omega_i & \text{otherwise}
\end{cases} \\
& V_{AFF} = V_i
\end{split}
\label{eq:AF_single_law}
\end{equation}
where $d_p$ represents the preceding distance at which a single peak was perceived in an earlier time-step of the navigation controller. Hence, only when the layer is activated consecutively will the single wall AFF behavior cause an output rotational velocity to align the mobile platform with this~wall. 

If two peaks are detected, full-corridor-following behavior is triggered. These peaks at lateral distances $d_l$ and $d_r$, respectively, need to be larger than the threshold $T_{AFF,corr}$ and should be on opposite sides of $d=0$. The~full-corridor-following behavior is calculated as
\begin{equation}
\begin{split}
& \omega_{AFF} = \omega_i + \lambda_{AFF} (d_l - d_r)\\
& V_{AFF} = V_i
\end{split}
\label{eq:AF_corridor_law}
\end{equation}

Gain factor $\lambda_{AFF}$ is the same single-wall and full-corridor-following behavior. It is set to 1 in simulation and 0.1 for the real experiments.
\section{Experimental~Results}\label{sec:results}
To validate the theoretical model for acoustic flow and the developed layered navigation controller, experiments were performed in both simulation as well as on a real mobile platform. This required both an offline and online version of the layered navigation controller. We will discuss these two validation techniques separately. 
\subsection{Simulation}
A simulation framework was made that included simulation of the indoor environment, robot, and sonar sensors. This framework was developed in MATLAB. By~using simulation, the~controller could be tested safely and support a numerical evaluation. The~simulation framework works offline and uses a fixed time-step of \SI{0.1}{\second}. The~previous version of this simulation framework used in~\cite{Steckel2017} was heavily extended to support multi-sonar, the~updated acoustic flow model, and the updated layered navigation controller using acoustic control regions. The simulation scenarios that were tested focused on navigating indoor environments with the primary goals to have no collisions and {reach the final waypoint without any short period of being stuck}. Furthermore, to~validate the multi-sonar paradigm, the~numerical analysis needed to demonstrate stability in the motion behavior regardless of the multi-sonar setup. The~simulation scenarios were designed with several corridors, doors, singular walls, and round obstacles. Additionally, two moving obstacles were added that would move in front of the mobile platform which could represent humans or other mobile~platforms. 

The layered navigation controller requires input velocities which, for the simulation scenarios, were given by a waypoint guidance system inspired by P. Boucher~\cite{Boucher2016}. This waypoint navigation is intended for differential steering vehicles such as the ones we use in the experiments. The~waypoints were placed sporadically and sparsely to ensure that the navigation controller was not unreasonably influenced by these input velocities, and subsequently calculated the local path to the next~waypoint.

The simulation of the eRTIS sensors required the simulation of the impulse responses of the reflections on the objects in the environment. We designed these responses for reflections such as corners, edges, and planar surfaces~\cite{Kuc1987}. Furthermore, occlusion of the sensors by objects is also simulated. Thereon, these impulse responses become the input of the time-domain simulation of the eRTIS sensor. In~this simulation, a virtualized array is used. The~same digital signal processing 
performs as a real eRTIS sensor does, as described in Figure~\ref{fig:ertis_processing}. The~outcome is a 2D energyscape of the horizontal plane in the frontal hemisphere of the eRTIS sensor. It has an FOV of \ang{180} which is divided during beamforming in steps of \ang{1}. The~maximum range is defined at  \SI{5}{\meter}. {Furthermore, within this post-processing pipeline, dead zones can be defined where the FOV of the sensor would be obscured by, for example, structures of the vehicle or other sonar sensors. In~the simulation experiments, the sensors were defined as rectangular objects of 12 cm by 5 cm for this FOV obstruction.}

\begin{figure}[H]
\begin{tabular}{cc}
\includegraphics[width=0.55\textwidth]{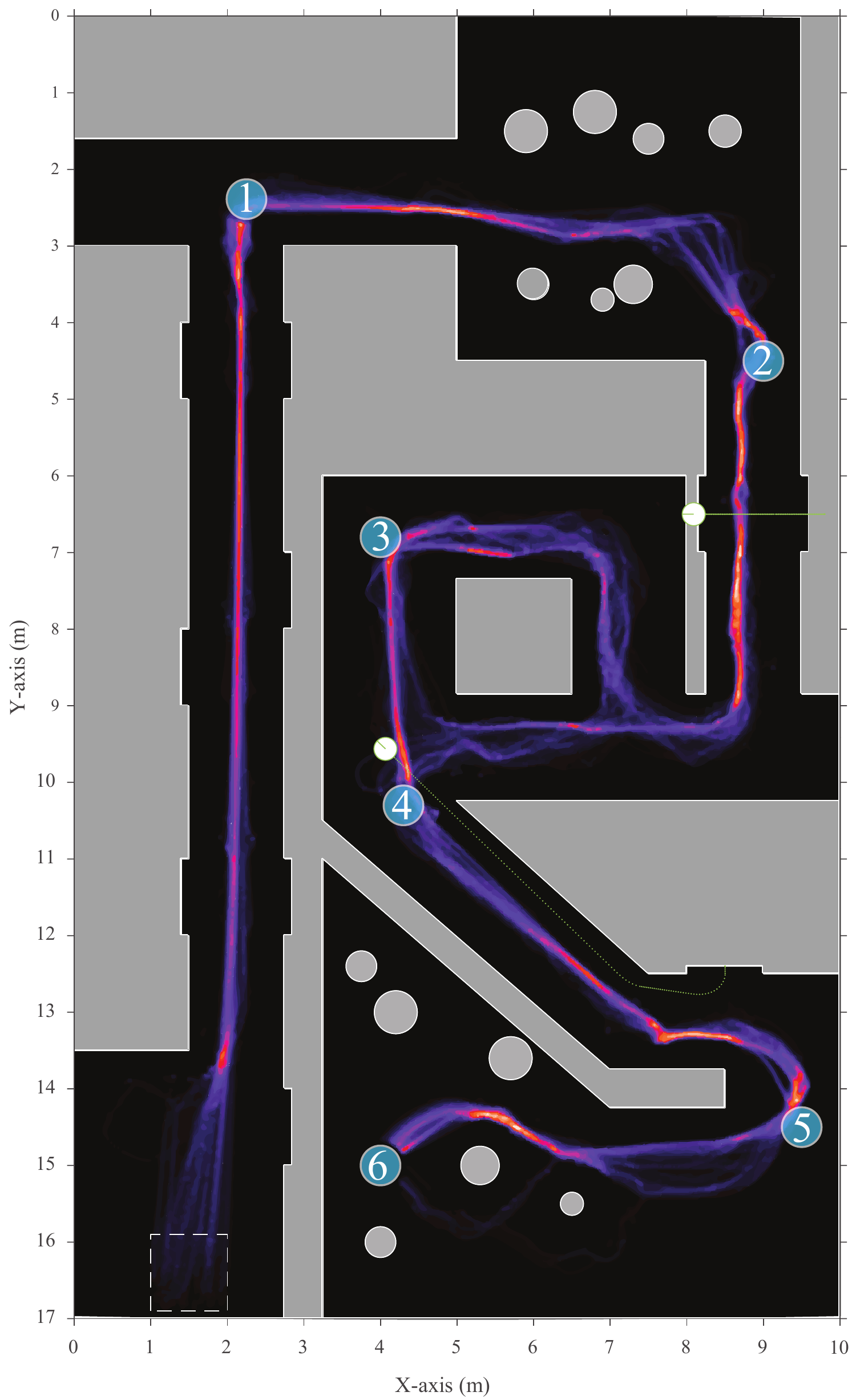}&\includegraphics[width=0.38\textwidth]{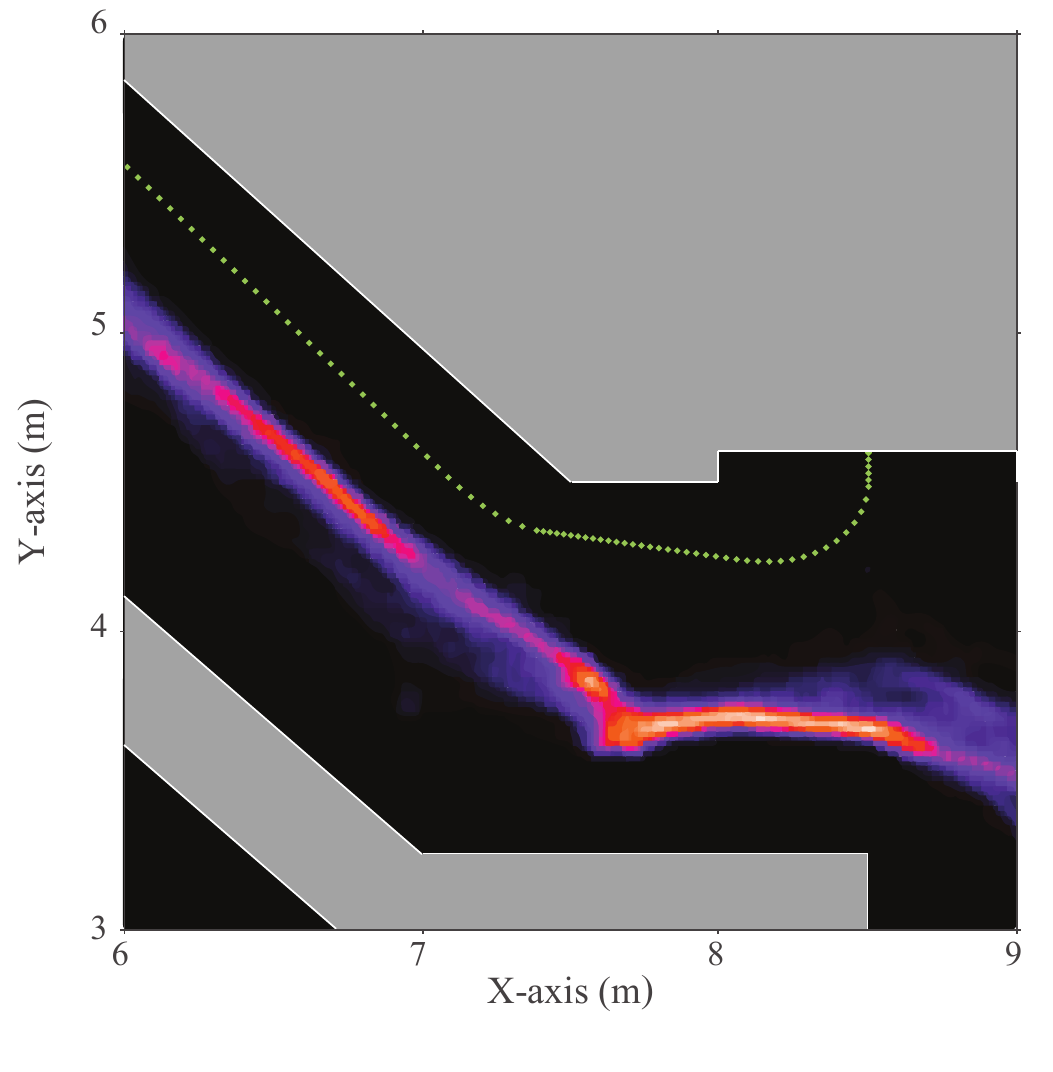}\\
(\textbf{a})&(\textbf{b})\\
\end{tabular}
  \caption{(\textbf{a}) A heat map of the trajectory distribution of the simulated scenario. It was taken over ten multi-sonar configurations. Each configuration was run 15 times for a total of 150 trajectories. The~start zone, indicated as a rectangle with dashed edges, was the location where the starting position is chosen at random. Afterwards, the~mobile platform would start moving between the sequential waypoints (shown as the blue numbered circles 1 to 6) based on the local layered navigation controller and global waypoint navigator. (\textbf{b}) A more detailed highlight of an area where a dynamic object interfered with the mobile platform path. This shows the adjustment the navigation controller made to avoid the dynamic object more~clearly.}
  \label{fig:results_simulation}
\end{figure}

The mobile platform is a circular robot with a diameter of \SI{20}{\cm}. It drives using differential steering up to a maximum velocity of \SI{0.3}{\meter/\second}. 
The modular design with support for multi-sonar of the layered navigation controller was validated by using 10~different multi-sonar configurations on this mobile platform. Each configuration has a single, two, or three simulated eRTIS devices, which are shown in Table~\ref{table:results_sim_configurations}. Each setup was run through the environment 15 times. The~resulting distribution heat map is shown in Figure~\ref{fig:results_simulation}. As~can be observed in this heat map of the scenario, and from in the subsequent numerical evaluation, not a single multi-sonar configuration tested caused a collision or other unwanted behavior. Furthermore, the~expected behavior of collision avoidance with both static and dynamic obstacles, navigating junctions 
 and corridor-following walls, could be seen. The~main variations that could be noted between the multi-sonar configurations are in the selection of the local navigation trajectory near junctions. This is due to the difference in the energyscapes, which differed among the multi-sonar configurations based on their~locations.

\begin{table}[H]
\centering
\caption{Simulation multi-sonar~setups.}
\label{table:results_sim_configurations}
\newcolumntype{C}{>{\centering\arraybackslash}X}
\begin{tabularx}{\textwidth}{CCCCCCCCCC}
\toprule
\multirow{2}{*}{\vspace{-4pt} \textbf{Setup}} & \multicolumn{3}{c}{\textbf{Simulated Sonar 1}} & \multicolumn{3}{c}{\textbf{Simulated Sonar 2}} & \multicolumn{3}{c}{\textbf{Simulated Sonar 3}} \\ \cmidrule{2-10}
& \boldmath{$\alpha$} \textbf{($^\circ$)} & \boldmath{$\beta$} \textbf{($^\circ$)} & \boldmath{$l$} \textbf{(cm)} & \boldmath{$\alpha$} \textbf{($^\circ$)} & \boldmath{$\beta$} \textbf{($^\circ$)} & \boldmath{$l$} \textbf{(cm)} & \boldmath{$\alpha$} \textbf{($^\circ$)} & \boldmath{$\beta$} \textbf{($^\circ$)} & \boldmath{$l$} \textbf{(cm)} \\ \midrule
1  & 0 & 0 & 18 &  & unused &   &   & unused &   \\ 
2  & 0 & $-$20 & 14  &90 & $-$10 & 10  & $-$90 & $-$5 & 8  \\ 
3  & 90 & $-$20 & 10  &$-$90 & 20 & 10  & & unused &\\ 
4  & 0 & 0 & 12 &90 & 0 & 12  & $-$90 & 0 & 12  \\ 
5  & 45 & 0 & 4  &$-$135 & 0 & 4  &  & unused &  \\ 
6  & 0 & 0 & 10  &$-$180 & 0 & 0  &   & unused &  \\ 
7  & 0 & 20 & 6  &90 & 10 & 0  & $-$90 & 20 & 14  \\ 
8  & 0 & 0 & 0  &120 & $-$120 & 14  & $-$120 & 120 & 14  \\ 
9  & 180 & $-$180 &6  &  & unused &   &   & unused & \\ 
10 & 45 & $-$10 & 6  &$-$45 & 10 & 6  & $-$180 & 0 & 0  \\ 
\bottomrule
\end{tabularx}
\end{table}
\unskip

\subsection{Experimental Mobile~Platform}
\begin{figure}[H]
\begin{tabular}{cc}
\includegraphics[width=0.52\textwidth]{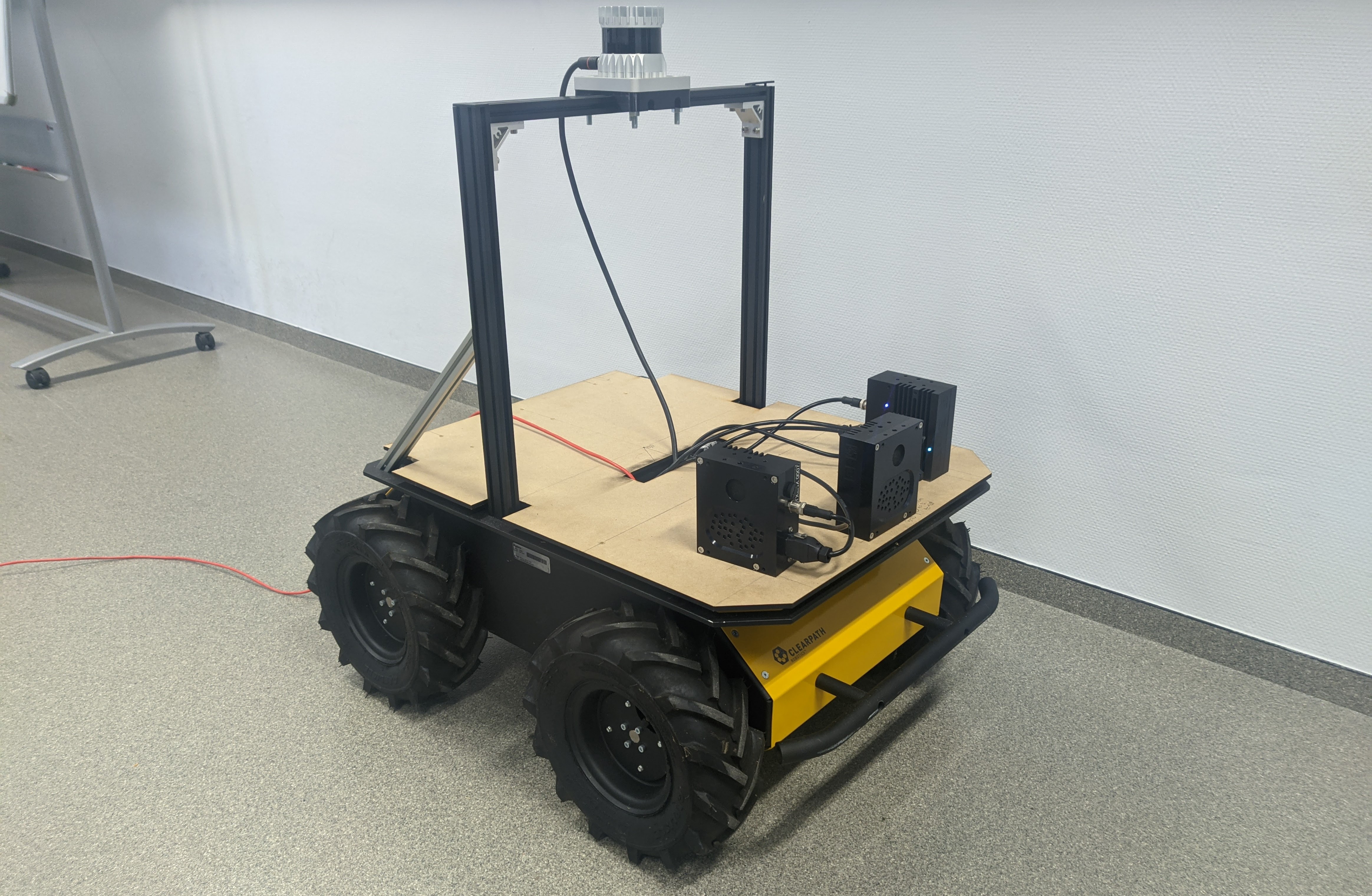}&\includegraphics[width=0.42\textwidth]{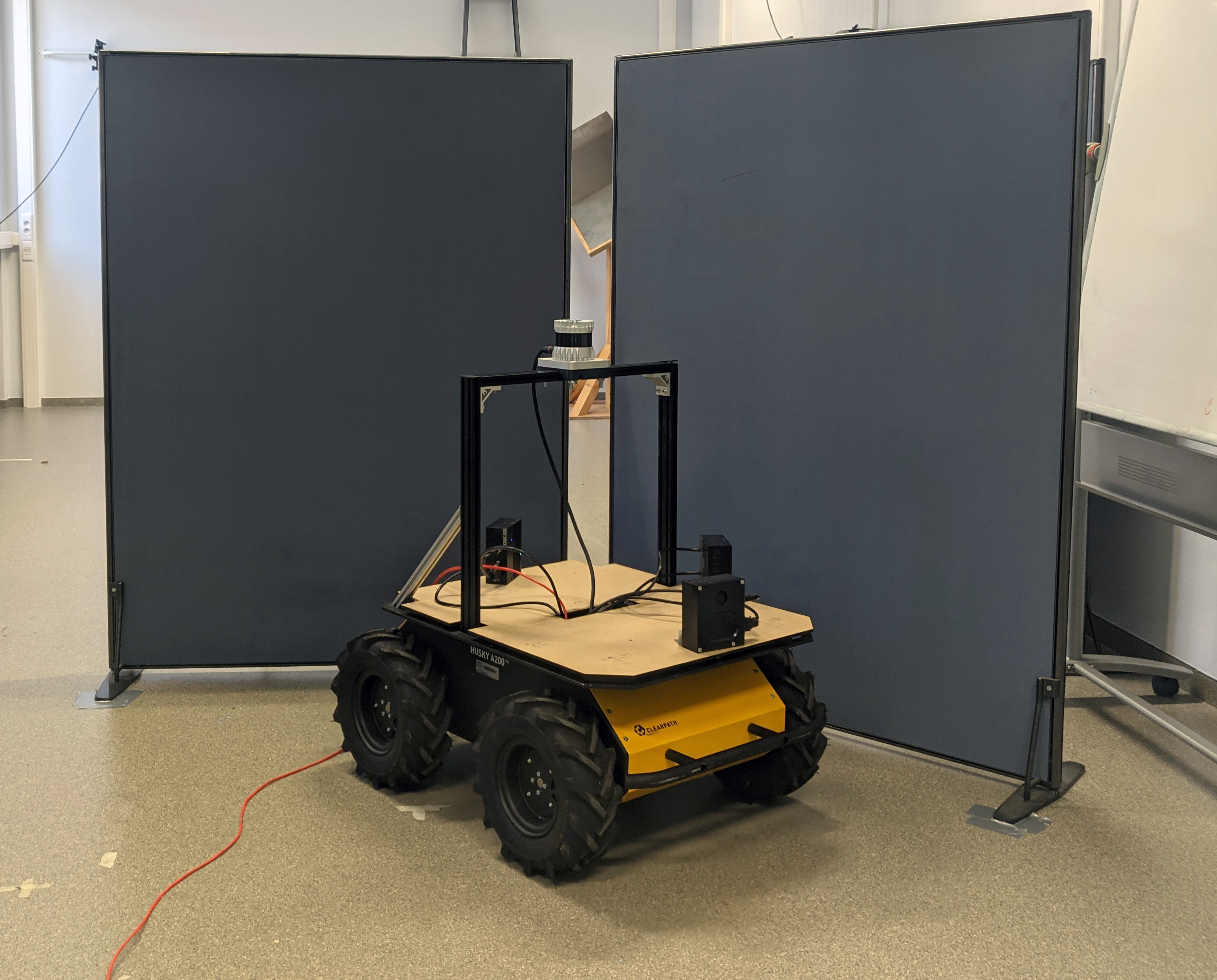}\\
(\textbf{a})&(\textbf{b})
\end{tabular}
  \caption{(\textbf{a}) The experimental setup with a Clearpath Husky UGV as mobile platform used for validating the controller behavior. This figure shows the setup with three mounted eRTIS sensors that each have a built-in NVIDIA Jetson TX2 NX module. Furthermore, an~Ouster OS0-128 LiDAR sensor is mounted on the top and is used for 2D map generation during the validation experiments. (\textbf{b}) A photo of the UGV taken during one of the experiments. More specifically, the~avoidance of a direct wall collision. {Subfigure (\textbf{a}) was moved from Figure~\ref{fig:sensor_images} to better fit the paper structure. Subfigure (\textbf{b}) was added to show one of the setups.}}
  \label{fig:vehicle_images}
\end{figure}

The exact same layered controller was translated from MATLAB to a real-time Python implementation for experiments using a real mobile platform. The~controller was connected to the software system described in~\cite{Jansen2020}. This system allows for sensor discovery, synchronized measurements, and real-time GPU-accelerated signal processing. The~mobile platform used was a Clearpath Husky UGV, as shown in Figure~\ref{fig:vehicle_images}a. {Equal to in the simulation, structures of the UGV, such as the pillars or sensors, obscuring each other's FOVs can be defined in the software system during the processing of the energyscapes. The~eRTIS sensors are defined by their real-life rectangular shape of 11.6 cm by 5 cm. The~2D energyscape of the real eRTIS sensors was the same as in the simulation, capturing the horizontal plane in the frontal hemisphere of the eRTIS sensor. It has an FOV of 180 degrees which is divided during beamforming in steps of 1 degree.} 



The Python implementation of the controller during these experiments was running on a i5-4570TE dual-core CPU in real time and could process the behavior layers for up to three eRTIS sensors at \SI{10}{\hertz}. The~average execution time of a single frame for the controller was \SI{40}{\milli\second} with a maximum of \SI{50.4}{\milli\second} during all experiments. Matching with the simulated experiments, one to three eRTIS sensors were used as they are described in Section~\ref{sec:ertis}. We tested six different multi-sonar configurations in several scenarios to test the behavior layers of the controller. These configurations are listed in Table~\ref{table:results_real_configurations}. Three scenarios were tested for each sensor configuration: avoiding an obstacle, avoiding a direct wall collision, and navigating a T-shaped~hallway.
\begin{table}[H]
\centering
\caption{Experimental mobile platform multi-sonar~setups.}
\label{table:results_real_configurations}
\newcolumntype{C}{>{\centering\arraybackslash}X}
\begin{tabularx}{\textwidth}{CCCCCCCCCC}
\toprule
\multirow{2}{*}{\vspace{-4pt} \textbf{Setup}} & \multicolumn{3}{c}{\textbf{Sensor 1}} & \multicolumn{3}{c}{\textbf{Sensor 2}} & \multicolumn{3}{c}{\textbf{Sensor 3}} \\ \cmidrule{2-10}
& \boldmath{$\alpha$} \textbf{($^\circ$)} & \boldmath{$\beta$} \textbf{($^\circ$)} & \boldmath{$l$} \textbf{(cm)} & \boldmath{$\alpha$} \textbf{($^\circ$)} & \boldmath{$\beta$} \textbf{($^\circ$)} & \boldmath{$l$} \textbf{(cm)} & \boldmath{$\alpha$} \textbf{($^\circ$)} & \boldmath{$\beta$} \textbf{($^\circ$)} & \boldmath{$l$} \textbf{(cm)} \\ \midrule
1  & 0 & 0 & 41  &30 & 60 & 40  & $-$30 & $-$60 & 40  \\ 
2  & 0 & 0 & 41  &180 & 0 & 27  & 58 & 32 & 31  \\ 
3  & 0 & $-$12 & 41  &180 & 35 & 27  & 58 & 2 & 31  \\ 
4  & 0 & 0 & 41 &  & unused &   &   & unused &   \\ 
5  & 140 & 80 & 25  &$-$20 & $-$30 & 30 & 30 & $-$30 & 39  \\ 
6  & 140 & 11 & 25  &$-$20 & 0 & 30 &   & unused &   \\ 
\bottomrule
\end{tabularx}
\end{table}

\begin{figure}[H]
\includegraphics[width=0.90\linewidth]{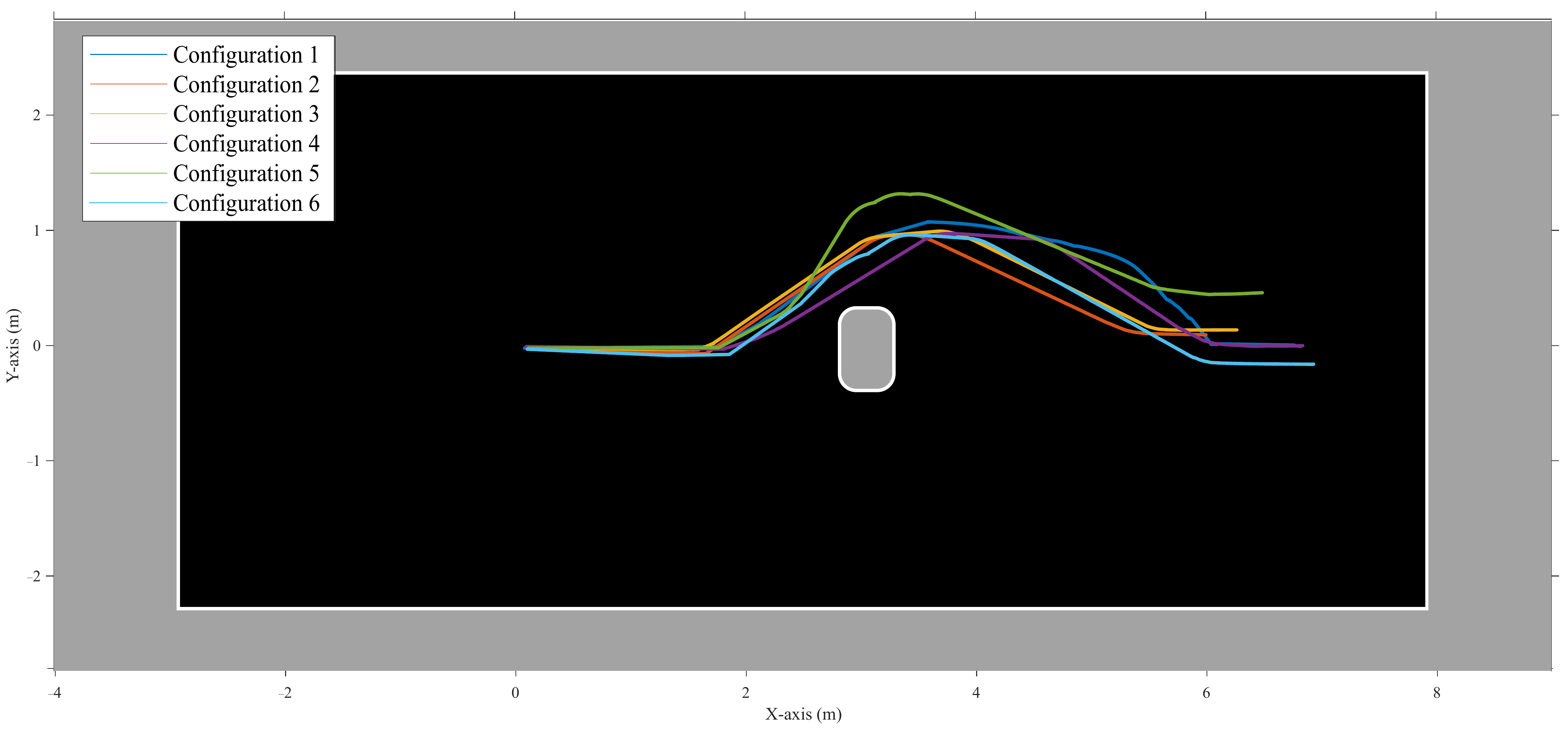}
\caption{Experimental results for a real mobile platform navigating a room with an obstacle that needs to be dealt with safely. The~room was empty except for the obstacle. The~figure shows the trajectories of the six different configurations that were used for validating the adaptability of the layered controller. The~trajectories were combined by using a single starting point. The~map shown here is a manually cleaned-up version of the occupancy grid results of a LiDAR SLAM algorithm that was run offline. This figure shows avoiding a static obstacle, a~chair in this instance. Although~the behavior layer focused on here is obstacle avoidance, all layers were active during the experiments. 
}
\label{fig:results_oa_all}
\end{figure}

\begin{figure}[H]
\begin{tabular}{cc}
 \includegraphics[width=0.85\textwidth]{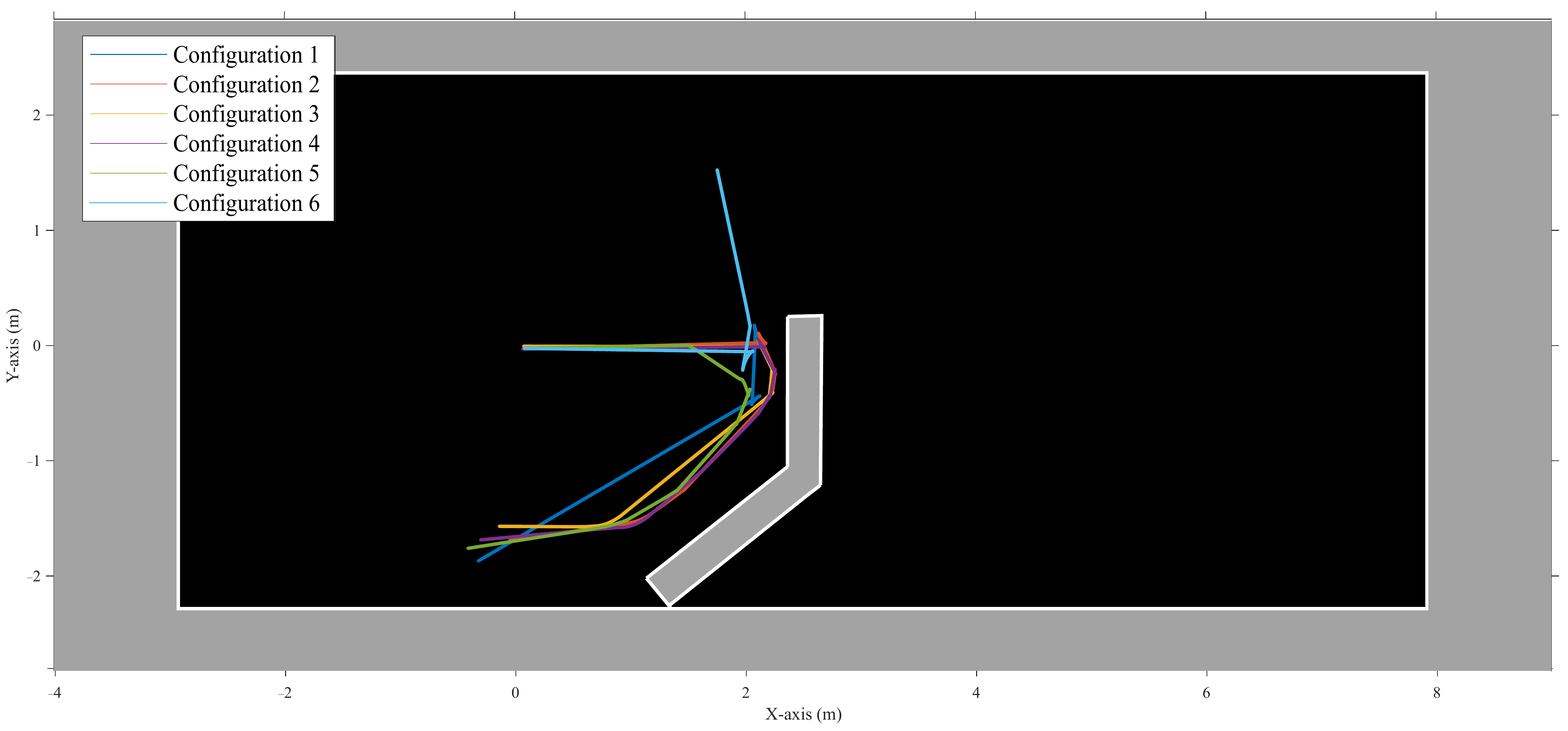}\\
 (\textbf{a})\\
 \includegraphics[width=0.85\textwidth]{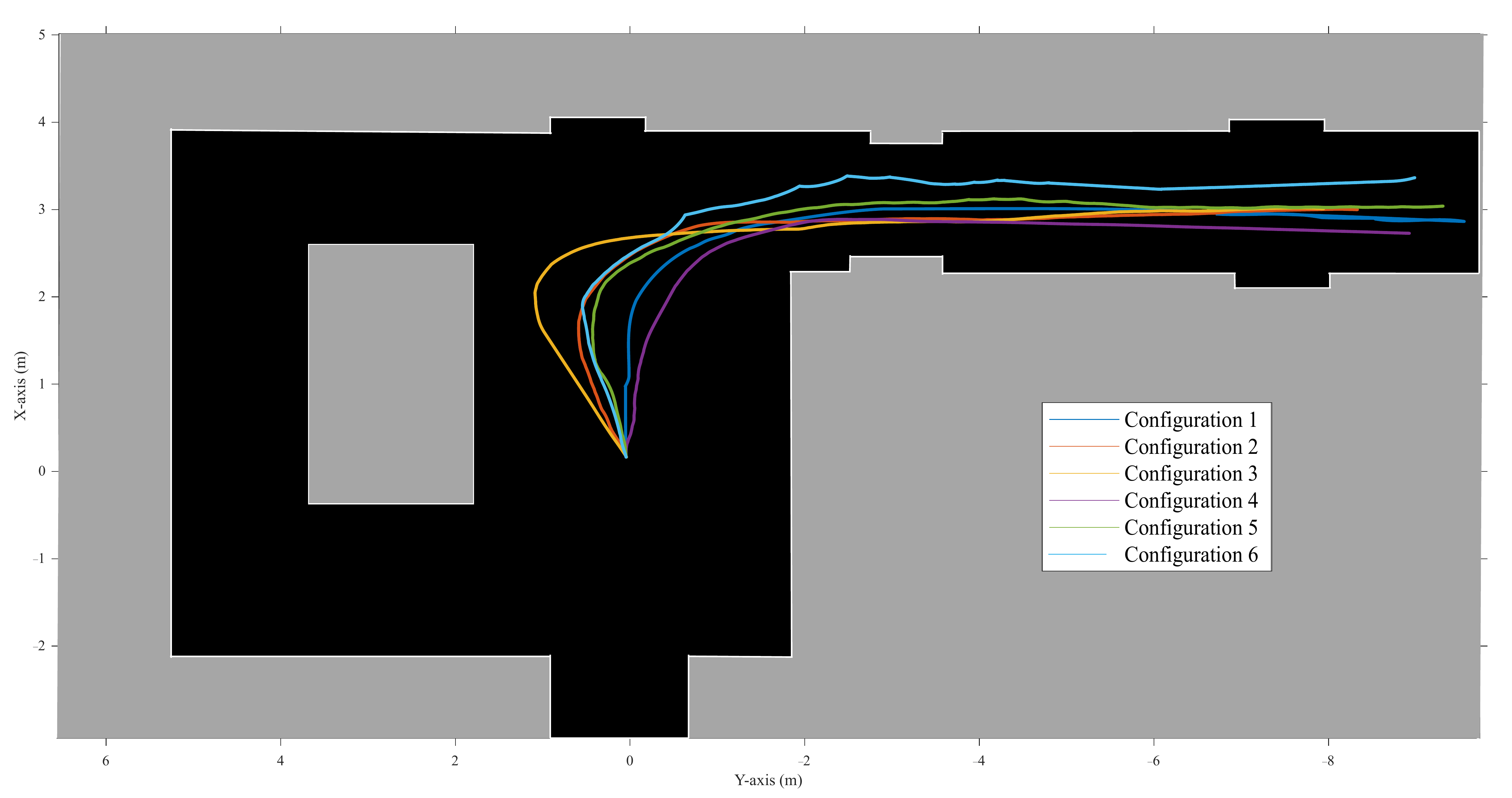}\\
 (\textbf{b})\\
 
  \includegraphics[width=0.85\textwidth]{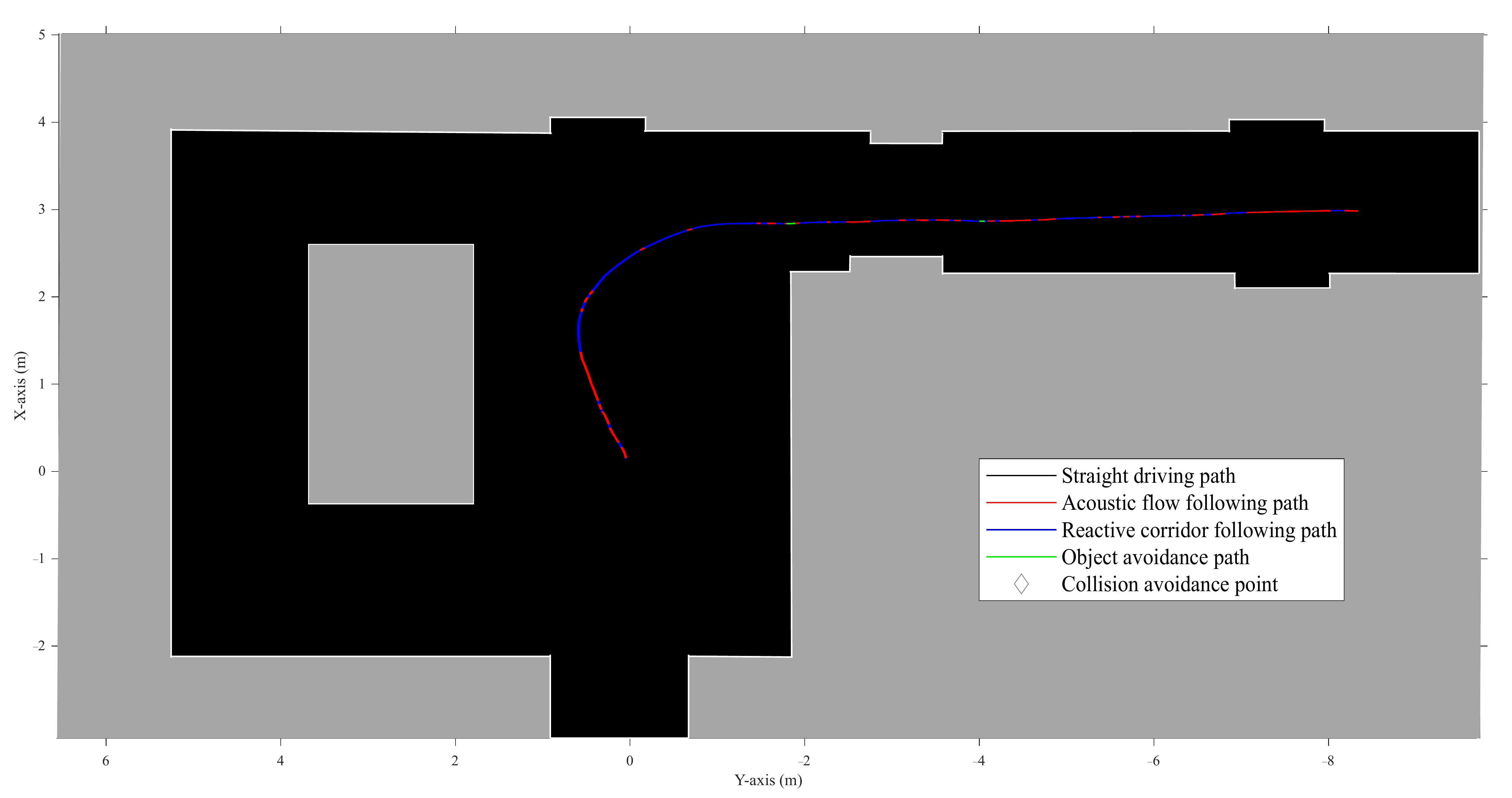}\\
  (\textbf{c})\\
  \end{tabular}
\caption{\textls[-25]{The maps shown are manually cleaned-up versions of the occupancy grid results of a LiDAR SLAM algorithm that was run offline. Although~the behavior layer focused on in each scenario is specific, all layers were active during the experiments. (\textbf{a}) Results for a real mobile platform navigating a room with an artificially added wall that needs to be dealt with safely. The~figure shows the trajectories of the six different configurations that were used for validating the adaptability of the layered controller. The~trajectories were combined by using a single starting point. (\textbf{b}) Experimental results for a real mobile platform navigating a T-shaped corridor. The~figure shows the trajectories of the six different configurations that were used for validating the adaptability of the layered controller. The~trajectories were combined by using a single starting point. (\textbf{c}) This figure shows one of the individual trajectories of (\textbf{b}), indicating for each point on the trajectory what layer of the controller was generating the output velocity.} 
}
\label{fig:results_ca_and_hallway}
\end{figure}

The motor velocity commands of the controller and resulting navigation trajectories for these scenarios can be seen in Figures~\ref{fig:results_oa_all} and \ref{fig:results_ca_and_hallway}. For~the real experiments, no waypoint navigation was applied, and the input linear velocity to the controller was static at \SI{0.3}{\meter/\second}. Figure~\ref{fig:vehicle_images}b shows the UGV during the experiment to avoid a direct wall collision.

However, for~two scenarios, a rotational input velocity was applied. Firstly, the~case of validating the obstacle avoidance layer seen in Figure~\ref{fig:results_oa_all}, when the platform had steered clear of the obstacle to go back to straight drive behavior afterwards. Similarly, when navigating the hallway, as seen in Figure~\ref{fig:results_ca_and_hallway}b,c, a~small rotational velocity was applied to steer the mobile platform to the right corridor instead of the left. The~same outcome could have been achieved by including a small bias in the behavior layer design for a specific direction, but~this was decided against for better observable validation of the~controller. 

An Ouster OS0-128 LiDAR was also mounted on the vehicle to create an occupancy-grid map of the environment with an offline implementation of Cartographer~\cite{Hess2016}. The~odometry of the mobile platform was taken from the wheel odometry of the Clearpath Husky UGV software. These occupancy maps were used to create a handmade recreation of the environment, as seen in Figures~\ref{fig:results_oa_all} and \ref{fig:results_ca_and_hallway}.

The focus of these experiments was to validate if the adaptive nature of the new behavior controller using the acoustic control regions would result in stable and safe navigational behavior for all configurations tested. From~these experiments, we could observe that the correct layer was activated for each combination of scenario and sensor layout configuration. No collisions could be observed. During~the collision avoidance experiment shown in Figure~\ref{fig:results_ca_and_hallway}a, one can observe that for a single configuration the controller decided to navigate to the opposite direction from the other five configurations. {However, at~no point did the robot become stuck for even a short period.} We find this to be expected, and in simulation it was also observed that not all configurations would handle each scenario in a similar way. Small differences in approach angle and resulting difference in acoustic energy on both sides of the mobile platform can result in a different choice of rotational velocity direction for all controller layers.\vspace{-6pt}

\unskip

\section{Conclusions and Future~Work}\label{sec:conclusion}
The various experiments in both simulation and real-wold scenarios validate the layered navigation controller for an autonomous mobile platform for safe {2D} navigation in spatially diverse environments featuring dynamic objects. The~controller can do so while supporting several sensors to be placed where possible or required. No collisions or {short periods of the mobile platform being stuck} were observed during the experiments. 
A significant benefit of this work is that spatial navigation is accomplished without the necessity for explicit spatial segmentation. Furthermore, a~user of this controller can choose their acoustic control regions to achieve specific behavior. 
The implemented primitive behavior layers are ample for safe motion between the waypoints in simulation or free-roam navigation, as shown in the real-world experiments. However, one could insert new layers for more specific motion behavior. For~example, when moving through a corridor, to always keep the mobile platform on the right side, or~to perform specific actions in cases where predefined objects such as for doors, charging stations, intersections, or elevators are detected. However, most of these suggestions would require some form of semantic knowledge about the environment to be known a priori or generated online. Nevertheless, the~simplicity of the subsumption layered architecture allows more layers to be added with simple and complex navigation~behaviors.

The real-time implementation on a real-experimental platform for both the signal processing and the behavior controller itself allows this system to be implemented on relatively low-end hardware with minimal computing power. This proves that it can be practically implemented for real industrial use-cases already without any additional~change. 

In the future, a~primary extension to the current implementation would be a comprehensive system for path planning and global navigation. Consequently, the~work presented here could provide the local navigation. Nevertheless, to~accomplish a more global navigation, more spatial knowledge of the environment is necessary. This knowledge can be generated by SLAM~\cite{Steckel2013b, Krekovic2016} or landmark beacons~\cite{Simon2020}.
\section{Materials and~Methods}\label{sec:materials}
The offline controller and the simulation were carried out in MATLAB 2021b using the following toolboxes: Image Processing Toolbox, Instrument Control Toolbox, Parallel Computing Toolbox, Phased Array System Toolbox, DSP System Toolbox, Signal Processing Toolbox, and Statistics and Machine Learning Toolbox. The~online implementation of the controller was created in Python v3.7, combined with the back-end system developed in~\cite{Jansen2020}. The~occupancy grids for validation of the experimental scenarios were made with Cartographer v2.0~\cite{Hess2016} based on the point-clouds of an Ouster OS0-128 LiDAR. ROS middleware~\cite{Quigley2009ROS:System} was used for recording the data packages of the behavior controller, wheel odometry, and LiDAR point-clouds.

\vspace{6pt} 
\authorcontributions{Conceptualization, W.J., D.L. and J.S.; methodology, W.J., D.L. and J.S.; software, W.J. and J.S.; validation, W.J.; investigation, W.J. and J.S.; writing---original draft preparation, W.J.; writing---review and editing, J.S.; visualization, W.J.; supervision, D.L. and J.S.; project administration, J.S. All authors have read and agreed to the published version of the~manuscript.}

\funding{This research received no external~funding.}

\institutionalreview{Not applicable.}

\informedconsent{Not applicable.}

\dataavailability{Data sharing not applicable. Experimental recordings and metrics are available upon request.} 

\conflictsofinterest{The authors declare no conflict of~interest.} 
 




\pagebreak
\begin{adjustwidth}{-\extralength}{0cm}

\reftitle{References}

\end{adjustwidth}
\end{document}